\documentclass[9.5bp,journal,compsoc]{IEEEtran}
\usepackage{epsfig}
\usepackage{graphicx}
\usepackage[pagebackref=false,breaklinks=true,colorlinks,linkcolor={green},citecolor={green},urlcolor={blue},bookmarks=false]{hyperref}
\usepackage{cite}
\usepackage{microtype}
\usepackage{subfigure}
\usepackage{url}

\usepackage{array}
\usepackage[normalem]{ulem}
\usepackage{ulem}
\usepackage{amsmath}
\usepackage{graphicx,subfigure}
\usepackage{longtable}
\usepackage{rotating}
\usepackage{multirow}
\usepackage{makecell}
\usepackage{bm}
\usepackage{float}
\usepackage{flushend}
\usepackage{xcolor}
\usepackage{soul}
\usepackage{booktabs}
\usepackage{threeparttable}
\usepackage{color}
\usepackage{algorithmic}
\usepackage{tabularx}
\usepackage{amsfonts}
\usepackage{epstopdf}
\usepackage{latexsym}
\usepackage{amssymb}
\usepackage{footnote}
\usepackage{pifont}
\usepackage[ruled]{algorithm2e}

\newtheorem{lemma}{Lemma}
\newtheorem{proof}{Proof}

\renewcommand{\IEEEQED}{\IEEEQEDopen}

\usepackage{threeparttable}

\begin{document}

\title{Distance-based  Fuzzy K-means Clustering without Cluster Centroids} %without Cluster Centroids

\author{Yichen~Bao, Han~Lu, Quanxue~Gao
%	 ~Xinbo~Gao,~\IEEEmembership{Senior Member,~IEEE}, and ~Dacheng~Tao,~\IEEEmembership{Fellow,~IEEE}
\IEEEcompsocitemizethanks{
\IEEEcompsocthanksitem This work was supported in part by the Natural Science Foundation of China under Grant 62176203, in part by the Fundamental Research Funds for the Central Universities and in part by the Innovation Fund of Xidian University(YJSJ24017). (Corresponding author: Quanxue Gao )\protect
%in part by Natural Science Basic Research Plan in Shaanxi Province (Grant 2020JZ-19)

\IEEEcompsocthanksitem Y. Bao, H. Lu, and Q. Gao are with the School of Telecommunications Engineering, Xidian University, Xi'an 710071, China (e-mail: qxgao@xidian.edu.cn; ).\protect

}%
%\thanks{Manuscript received XXXX; revised XXXX; accepted XXXX.}
}

\markboth{IEEE TRANSACTIONS on Fuzzy Systems}%
%\markboth{Journal of \LaTeX,~Vol.~XXX,~No.~XXX,~XXX~XXX}%
{Shell \MakeLowercase{\textit{Xia et al.}}: XXXX}

\IEEEtitleabstractindextext{%
\begin{abstract}
Fuzzy \textit{K}-Means clustering is a critical technique in unsupervised data analysis. Unlike traditional hard clustering algorithms such as \textit{K}-Means, it allows data points to belong to multiple clusters with varying degrees of membership, determined through iterative optimization to establish optimal cluster centers and memberships, thereby achieving fuzzy partitioning of data. However, the performance of popular Fuzzy \textit{K}-Means algorithms is sensitive to the selection of initial cluster centroids and is also affected by noise when updating mean cluster centroids. To address these challenges, this paper proposes a novel Fuzzy \textit{K}-Means clustering algorithm that entirely eliminates the reliance on cluster centroids, obtaining membership metrics solely through distance matrix computation. This innovation enhances flexibility in distance measurement between sample points, thus improving the algorithm's performance and robustness. The paper also establishes theoretical connections between the proposed model and popular Fuzzy \textit{K}-Means clustering techniques. Experimental results on several real datasets demonstrate the effectiveness of the algorithm.
\end{abstract}

% Note that keywords are not normally used for peer-reviewed papers.
\begin{IEEEkeywords}
Clustering, Fuzzy K-Means, unsupervised learning.
\end{IEEEkeywords}}

\maketitle

\IEEEdisplaynontitleabstractindextext
\IEEEpeerreviewmaketitle

\IEEEraisesectionheading{\section{Introduction}\label{sec:introduction}}
\IEEEPARstart{C}{lustering} algorithms are a pivotal component of unsupervised learning methods in data mining and machine learning. Their primary objective is to partition a dataset of unlabeled data into multiple distinct clusters, reflecting either the inherent or global structure of the data. Clustering is instrumental in revealing data distribution patterns, identifying outliers, compressing data, and offering foundational knowledge for subsequent supervised learning tasks. These algorithms have been extensively applied across various fields such as market analysis, image processing, bioinformatics, and social network analysis, establishing themselves as a key tool in solving real-world problems.

Numerous clustering algorithms have been developed, including spectral clustering~\cite{NgJW01,YangGXYG22}, \textit{K}-Means~\cite{MacQueen1967,LuXWGYG24,NieLWL23}, Kernel \textit{K}-Means~\cite{KimLLL05,WangLLNL22,RenS021}, and Fuzzy \textit{K}-Means~\cite{DunnFCM1973,XuHXN16,NieZWLL22,WangZNL23}, among others.

Spectral clustering~\cite{NgJW01}, a graph-based clustering method~\cite{pami/ShiM00}, transforms clustering into an optimal graph partitioning problem. It begins by decomposing the graph's Laplacian matrix to derive a low-dimensional sample representation, followed by employing \textit{K}-Means to segment data points into \textit{K} clusters. Similar to other graph-based clustering methods, spectral clustering involves constructing a similarity adjacency matrix among samples, performing eigendecomposition on this matrix, and then clustering the resulting low-dimensional representation. Although graph-based approaches often yield superior results, constructing the adjacency matrix and the instability of eigendecomposition are notable drawbacks.

In contrast, \textit{K}-Means clustering~\cite{MacQueen1967} is renowned for its simplicity and has been a staple method since its inception. This algorithm partitions a dataset into \textit{K} subsets, iteratively updating cluster centroids and reassigning clusters to minimize the sum of distances between samples and their respective centroids, thereby achieving distinct and compact clustering. Extensions of \textit{K}-Means, like \textit{K}-Means++~\cite{ArthurV07}, address the randomness in initial centroid selection. \textit{K}-Means++ strategically selects initial centroids to enhance the likelihood of choosing representative sample points. For datasets not amenable to linear separation, Kernel \textit{K}-Means~\cite{KimLLL05,WangLLNL22} employs kernel methods to map data into a high-dimensional space, facilitating linear separability. The choice of kernel function in this context is critical and should reflect the data characteristics and problem complexity. Multi-kernel \textit{K}-Means~\cite{RenS021} integrates multiple kernel functions, proving effective in clustering complex, high-dimensional data. Additionally, \textit{K}-Medoids~\cite{RenHC22,NewlingF17} offers a robust alternative to traditional \textit{K}-Means, using the most centrally located object in a cluster as the cluster center, thereby reducing the influence of outliers.

% Despite the decent performance achieved by these extension algorithms of \textit{K}-Means, they still require the computation of cluster centroids, resulting in unstable clustering performance.

The spectral clustering and \textit{K}-Means clustering mentioned above, along with their related extension algorithms, are categorized as hard clustering. This approach strictly assigns each object to a specific cluster, following an "either-or" principle. However, many real-world datasets exhibit fuzzy rather than clear category boundaries, a scenario where hard clustering algorithms struggle due to their inability to handle such fuzziness. To address this, the Fuzzy \textit{K}-Means clustering algorithm~\cite{DunnFCM1973, XuHXN16, NieZWLL22, WangZNL23} was introduced. It incorporates fuzzy theory to determine the degree of membership of each sample to all cluster centers by optimizing the objective function. This method's strength lies in its capacity to express intermediate sample memberships, thereby capturing the uncertainty inherent in sample categorization and more accurately reflecting real-world scenarios. Enhanced versions of the original Fuzzy \textit{K}-Means clustering have been developed, such as the Maximum Entropy Fuzzy Clustering algorithm~\cite{ZhouCCZL16, LiNCH08}, which constructs a bipartite graph based on information entropy. Furthermore, Par et al. introduced a hybrid Possibilistic Fuzzy \textit{K}-Means algorithm~\cite{ZhouCCZL16, tang2019possibilistic}, combining Fuzzy \textit{K}-Means with Possibilistic \textit{K}-Means and applying a more relaxed constraint on the membership matrix.

Despite these advancements, the algorithmic process of Fuzzy \textit{K}-Means clustering mirrors that of traditional \textit{K}-Means, requiring random initialization of cluster centroids and their update through weighted mean calculation. This dependency significantly impacts the performance and robustness of clustering results. Attempts to eliminate the influence of cluster centers in \textit{K}-Means clustering have led to innovations like Ding's~\cite{DingH05} proof of the equivalence of non-negative matrix factorization to kernel \textit{K}-Means clustering with relaxed orthogonal relationships. This method calculates clustering labels through matrix factorization. Additionally, connections between kernel \textit{K}-Means and spectral clustering have been established~\cite{DhillonGK04}. However, these methods necessitate post-processing, which can compromise clustering quality and deviate from the original problem's solution. Centerless clustering, as proposed by Pei et al.~\cite{PeiC00023}, minimizes the sum of intra-cluster distances without considering cluster centroids. Yet, its assumption of equal sample numbers in each cluster limits its effectiveness on real-world datasets with cluster imbalances. Currently, no research directly addresses clustering without considering cluster centroids in the context of Fuzzy \textit{K}-Means.

This paper aims to introduce a centroid-less Fuzzy \textit{K}-Means clustering algorithm. Our method neither requires the random selection of initial cluster centers nor their subsequent updating, enhancing both the performance and robustness of the Fuzzy \textit{K}-Means algorithm. We focus on the principle of updating cluster centroids and incorporate this into the objective function's optimization, thereby obviating the need for considering cluster centroids. Our theoretical analysis confirms that our proposed model is equivalent to the standard Fuzzy \textit{K}-Means clustering algorithm. The primary contributions of our model are:
\begin{itemize}
	\item It completely eliminates reliance on cluster centroids, calculates the membership matrix directly using the distance matrix, and facilitates fuzzy partitioning of the dataset.
	\item It offers more flexible distance measurement between samples, expanding the applicability of the Fuzzy \textit{K}-Means clustering algorithm.
\end{itemize}

\textbf{\textit{Notaions:}} Throughout the paper, bold uppercase letters, such as $\textbf{A}$, represent metrics, and bold lowercase letters, such as $\textbf{a}$, denote vectors. $\textbf{a}_i$ denotes the $i$-th row of $\textbf{A}$ and $\textbf{a}_{:j}$ denotes the $j$-th column of $\textbf{A}$. Lowercase $a_{i}$ and uppercase $a_{ij}$ respectively denotes the elements of vector $\textbf{a}$ and matrix $\textbf{A}$. The Frobenius norm is denoted by $\|\bullet\|_F$, and $\textrm{tr}(\bullet)$ denotes the trace of matrix. The set of cluster index metrics is represented by "$Ind$".

% \begin{table}
	% \caption{Summary of notations}\label{notation}
	% \begin{tabularx}{1.0\linewidth} {
			% 	l
			% 	>{\raggedright\arraybackslash}X
			% 	}
		% 	\hline
		% 	Notation & Description \\ \hline
		% 	$\bm{\mathcal{X}}$ & Multi-modal data \\
		% 	$\textbf{X}^{\textrm{({\emph{v}})}}$ & The feature matrix in the \emph{v}-th modal \\
		% 	\emph{N} & The number of samples \\
		% 	\emph{K} & The number of clusters \\
		% 	\emph{V} & The number of modalities \\
		
		% 	$\textbf{A}^{\textrm{({\emph{v}})}}$ & The adjacency matrix in the \emph{v}-th modal \\
		% 	$\mathcal{L}_{\textrm{Tclu}}$ & The loss of clustering \\
		% 	$\mathcal{L}_{\textrm{Tcon}}$ & The contrastive loss \\
		% 	$\bm{\varpi}$ & The parametric model \\
		% 	\hline
		% \end{tabularx}
	% \end{table}

\section{Related Works}

In this section, we introduce the related works including \textit{K}-Means and Fuzzy \textit{K}-Means.

\subsection{\textit{K}-Means Clustering}

\textit{K}-Means clustering~\cite{MacQueen1967} partitions a given sample set. This method segregates the data set into \textit{K} distinct subsets, with each subset representing a cluster. Given an input data matrix $\mathbf{X} \in \mathbb{R}^{N \times d}$, where $N$ denotes the number of samples and $d$ represents the dimensionality of each sample, the goal of \textit{K}-Means clustering is to minimize the following objective function:
\begin{equation}\label{k-means1}
	\begin{aligned}
		\mathop {\min }\limits_{  \textbf{X}_1,\ldots, \textbf{X}_K}{ \sum\limits_{l=1}^{K}\sum\limits_{\textbf{x}_i\in  \textbf{X}_l}\left\|\textbf{x}_i - \textbf{u}_l\right\|_2^2}
	\end{aligned}
\end{equation}
where $K$ denotes the number of clusters in the data set, and $\textbf{X}_l$ represents the set of samples that belong to the $l$-th cluster. The centroid of the $l$-th cluster, denoted by $\textbf{u}_l$, is calculated as the mean of all samples in $\textbf{X}_l$. This is expressed by the formula
\[
\textbf{u}_l = \frac{1}{|\textbf{X}_l|} \sum_{\textbf{x}_i \in \textbf{X}_l} \textbf{x}_i,
\]
where $|\textbf{X}_l|$ denotes the number of samples in the set $\textbf{X}_l$.

If $\textbf{F} \in \text{Ind}$ represents the cluster indicator matrix, then for the $i$-th sample $\textbf{x}_i$, the element $f_{ij}$ of $\textbf{F}$ is defined as follows: $f_{ij}=1$ if $\textbf{x}_i$ belongs to cluster $\textbf{X}_j$, and $f_{ij}=0$ otherwise. Given this definition, equation~(\ref{k-means1}) can be reformulated as:
\begin{equation}\label{k-means1}
	\begin{aligned}
		\mathop {\min }\limits_{\textbf{F}\in Ind, \textbf{U}}{ \sum\limits_{i=1}^{N}\sum\limits_{l=1}^{K} f_{il}\left\|\textbf{x}_i - \textbf{u}_l\right\|_2^2}
	\end{aligned}
\end{equation}

The \textit{K}-Means clustering algorithm begins by randomly selecting $K$ initial centroids. Each data point is then assigned to its nearest centroid, effectively forming clusters. This assignment aims to minimize the objective function given in Eq.~(\ref{k-means1}). After assigning the points, the centroids of each cluster are recalculated. This process of assignment and recalibration is repeated until the centroids stabilize (i.e., no longer change) or a predetermined number of iterations is reached. Ultimately, the algorithm partitions the data into $K$ distinct clusters.

\subsection{Fuzzy \textit{K}-Means Clustering}

In \textit{K}-Means clustering, each sample is assigned to exactly one cluster, a method known as hard clustering. In contrast, Fuzzy \textit{K}-Means clustering introduces a concept of fuzzy ``membership'', allowing samples to be associated with multiple clusters to varying degrees. This approach, referred to as soft clustering, enables Fuzzy \textit{K}-Means to yield more nuanced clustering results, particularly in complex datasets where data points may overlap.

\subsubsection{\textbf{Fuzzy \textit{K}-Means Clustering}}
Fuzzy \textit{K}-Means clustering~\cite{DunnFCM1973} extends the traditional \textit{K}-Means clustering algorithm by introducing fuzzy parameters. In Fuzzy \textit{K}-Means, each data point can belong to multiple cluster centroids, rather than being strictly assigned to a single cluster centroid as in traditional hard clustering. This inclusion of fuzzy parameters allows Fuzzy \textit{K}-Means to more effectively handle uncertainties and fuzziness in data, thus offering a more accurate reflection of the inherent structure and relationships within the dataset.

Let $\textbf{Y}$ denote the membership matrix, where the element $y_{ij}$, ranging from $0$ to $1$, represents the degree of membership of the $i$-th sample to the $j$-th cluster. Given input data $\textbf{X}\in \mathbb{R}^{N\times d}$, the objective function of the standard Fuzzy \textit{K}-Means clustering is defined as:
\begin{equation}\label{stdfcm}
	\begin{aligned}
		\mathop {\min }\limits_{\textbf{Y}, \textbf{U}}&{ \sum\limits_{i=1}^{N}\sum\limits_{l=1}^{K} y_{il}^{r}\left\|\textbf{x}_i - \textbf{u}_l\right\|_2^2} \\
		&s.t. \quad \textbf{Y1}=\textbf{1}, \textbf{Y}\ge\textbf{0}
	\end{aligned}
\end{equation}
By employing the Lagrange multiplier method, we aim to determine the optimal solutions for the variables $\mathbf{Y}$ and $\mathbf{U}$. The formulation of the function to be solved is presented below:
\begin{equation}\label{stdfcm_y}
	\begin{aligned}
		y_{il} = \frac{\left\|\textbf{x}_i - \textbf{u}_l\right\|_2^{-\frac{2}{r-1}}}{\sum\limits_{j=1}^{K} \left(\left\|\textbf{x}_i - \textbf{u}_j\right\|_2^{-\frac{2}{r-1}}\right)}\\
	\end{aligned}
\end{equation}
\begin{equation}\label{stdfcm_u}
	\begin{aligned}
		\textbf{u}_l = \frac{\sum\limits_{i = 1}^{N}u_{il}^r \textbf{x}_i}{\sum\limits_{i = 1}^{N}u_{il}^r}
	\end{aligned}
\end{equation}

The Fuzzy \textit{K}-Means clustering algorithm also requires the selection of \textit{K} initial cluster centroids. It updates the membership degree matrix and cluster centroids iteratively. The stopping criteria for iterative FCM algorithms is given by $\max \left| y_{il}^{(t)} - y_{il}^{(t-1)} \right| < \epsilon$, where $\epsilon$ is a small positive number, and $t$ denotes the number of iterations~\cite{Bezdek81}.

%The Fuzzy \textit{K}-Means clustering algorithm also requires the selection of \textit{K} initial cluster centroids. For each sample point, the membership degree of each cluster is calculated based on the distance between that point and the cluster centroid. Then, the centroids of each cluster are calculated based on the current membership degree values involving a weighted average of data points, with weights determined by the membership values. Subsequently, the membership matrix and cluster centers are iteratively updated until the algorithm converges.
%In each iteration, the membership degrees and cluster centers of sample points are adjusted to better reflect the relationships between data points and clusters.

\subsubsection{\textbf{Robust and Sparse Fuzzy \textit{K}-Means Clustering}}

The performance of Fuzzy \textit{K}-Means can be enhanced through regularization techniques~\cite{5584538}. Xu et al. proposed an advanced variant, the Robust and Sparse Fuzzy \textit{K}-Means clustering (RSFKM)~\cite{XuHXN16}, which integrates robustness and sparsity into the clustering process. The objective function of RSFKM is given by:
\begin{equation}\label{rsfkm}
	\begin{aligned}
		\mathop {\min }\limits_{\textbf{Y}, \textbf{U}}&{ \sum\limits_{i=1}^{N}\sum\limits_{l=1}^{K} y_{il}\hat{d}_{il} + \lambda \|\textbf{Y}\|_F^2}\\
		&s.t. \quad \textbf{Y1}=\textbf{1}, \textbf{Y}\ge\textbf{0}
	\end{aligned}
\end{equation}
where $\hat{d}_{il}$ is a measure between $\textbf{x}_i$ and $\textbf{u}_l$, e.g. $\hat{d}_{il} = \left\|\textbf{x}_i - \textbf{u}_l\right\|_2^2$ is commonly used. The RSFKM define $\hat{d}_{il}$ as $\hat{d}_{il} = \left\|\textbf{x}_i - \textbf{u}_l\right\|_2$ ($\ell_{2,1}$-norm) and  $\hat{d}_{il} = min(\left\|\textbf{x}_i - \textbf{u}_l\right\|_2, \epsilon)$ (capped  $\ell_{1}$-norm), where $\epsilon$ is a threshold.

Suppose that $\hat{d}_{il}$ in Eq.~\eqref{rsfkm} represents the squared Euclidean distance. Then, Eq.~\eqref{rsfkm} can be expressed as:
\begin{equation}\label{rsfkm_euc}
	\begin{aligned}
		\mathop {\min }\limits_{\textbf{Y}, \textbf{U}}&{ \sum\limits_{i=1}^{N}\sum\limits_{l=1}^{K} y_{il}\left\|\textbf{x}_i - \textbf{u}_l\right\|_2^2 + \lambda \|\textbf{Y}\|_F^2}\\
		&s.t. \quad \textbf{Y1}=\textbf{1}, \textbf{Y}\ge\textbf{0}
	\end{aligned}
\end{equation}

\section{Methodology}

\subsection{Problem Formulation and objective}
Fuzzy \textit{K}-Means clustering is a widely utilized soft clustering algorithm. Unlike traditional \textit{K}-Means clustering and other hard clustering methods, Fuzzy \textit{K}-Means offers distinct advantages. It permits data points to be part of multiple clusters, each with varying degrees of membership. This flexibility makes it more adept at handling overlapping data, delivering more nuanced clustering outcomes. However, akin to \textit{K}-Means, it confronts challenges, notably sensitivity to initial cluster centroid selection and vulnerability to outliers, owing to its reliance on centroid computation.
To address these issues, we introduce Lemma~\ref{lemma1}, which re-expresses Fuzzy \textit{K}-Means from the perspective of manifold learning.
	\begin{lemma}\label{lemma1}
		Let $d_{ij} = \left\|\textbf{x}_i -\textbf{x}_j\right\|_2^2$ is the distance, $\textbf{Y}$ is the membership matrix and $\sum_j y_{ij}= 1$, $y_{ij} \ge 0$, then the following formula holds
	\end{lemma}
	\begin{equation}\label{le1_main}
		\begin{aligned}
			\mathop {\min }\limits_{\textbf{Y}, \textbf{U}}\sum\limits_{i}\sum\limits_{j} y_{ij}\left\|\textbf{x}_i - \textbf{u}_j\right\|_2^2 \Leftrightarrow   \mathop {\min }\limits_{\textbf{Y}}s_{ij}\|\textbf{x}_i - \textbf{x}_j\|_2^2
		\end{aligned}
	\end{equation}
	where the manifold structure $\textbf{S} = \textbf{Y}\textbf{P}^{-1}\textbf{Y}^\textrm{T}$ represents the cluster structure in the data. $\textbf{P}$ is a diagonal matrix and $p_{jj} = \hat{p}_j = \sum\limits_i y_{ij}$.

\begin{proof}
	Let $J = \sum\limits_i\sum\limits_j y_{ij}\left\|\textbf{x}_i - \textbf{u}_j\right\|_2^2$, and then $\frac{\partial J}{\partial\textbf{u}_j} = 2\sum\limits_i y_{ij} \textbf{u}_j - 2\sum\limits_i \textbf{x}_i y_{ij} $. Then $\textbf{u}_j$ can be solved by taking $\frac{\partial J}{\partial\textbf{u}_j} = 0$ because $J$ is a convex function. Thus we have the following
	\begin{equation}\label{pf1_u_j}
		\begin{aligned}
			\textbf{u}_j = \frac{\sum\limits_i \textbf{x}_i y_{ij}}{\sum\limits_i y_{ij}} = \textbf{y}_{:j}^\textrm{T}\textbf{X}\hat{p}_j^{-1}
		\end{aligned}
	\end{equation}
	where $\hat{p}_j = \sum\limits_i y_{ij}$ and $\textbf{y}_{:j}$ is the $j$-th column of $\textbf{Y}$. Substituting Eq.~(\ref{pf1_u_j}) into $J$, and then it is clear that
	\begin{equation}\label{pf1_J}
		\begin{aligned}
		&\sum\limits_i\sum\limits_j y_{ij}\left\|\textbf{x}_i - \textbf{u}_j\right\|_2^2 \\
			&\quad\quad= \sum\limits_i \textbf{x}_i\textbf{x}_i^\textrm{T}\sum\limits_j y_{ij} + \sum\limits_j \textbf{u}_j\textbf{u}^\textrm{T}_j\sum\limits_i y_{ij} - \\
			&\quad\quad\quad\quad 2\sum\limits_j \textbf{u}_j\sum\limits_i\textbf{x}_i^\textrm{T} y_{ij} \\
			&\quad\quad= \sum\limits_i \textbf{x}_i\textbf{x}_i^\textrm{T} + \sum\limits_j \textbf{y}_{:j}^\textrm{T}\textbf{X}\textbf{X}^\textrm{T}\textbf{y}_{:j}\hat{p}_j^{-1} - \\
			&\quad\quad\quad \quad 2\sum\limits_j \textbf{y}_{:j}^\textrm{T}\textbf{X}\textbf{X}^\textrm{T}\textbf{y}_{:j}\hat{p}_j^{-1} \\
			&\quad\quad=\sum\limits_i \textbf{x}_i\textbf{x}_i^\textrm{T} - \sum\limits_j \textbf{y}_{:j}^\textrm{T}\textbf{X}\textbf{X}^\textrm{T}\textbf{y}_{:j}\hat{p}_j^{-1} \\
			&\quad\quad= \textrm{tr}(\textbf{X}^\textrm{T}\textbf{X}) - \textrm{tr}(\textbf{X}^\textrm{T}\textbf{Y}\textbf{P}^{-1}\textbf{Y}^\textrm{T}\textbf{X})\\
			&\quad\quad= \textrm{tr}\left(\textbf{X}^\textrm{T}(\textbf{I} - \textbf{Y}\textbf{P}^{-1}\textbf{Y}^\textrm{T})\textbf{X}\right)
		\end{aligned}
	\end{equation}
	where $\textbf{P}$ is a diagonal matrix and $p_{jj} = \hat{p}_j = \sum\limits_i y_{ij}$. Let $\textbf{S} = \textbf{Y}\textbf{P}^{-1}\textbf{Y}^\textrm{T}$ and $\textbf{F} = \textbf{YP}^{-\frac{1}{2}}$, then it is easy to see $\textbf{S1} = \textbf{1}$ and $\textbf{S} = \textbf{F}\textbf{F}^\textrm{T}$. Thus, $J$ can be rewritten as follows:
	\begin{equation}\label{pf1_J}
		\begin{aligned}
			J &=\textrm{tr}\left(\textbf{X}^\textrm{T}(\textbf{I} - \textbf{S})\textbf{X}\right)= \sum\limits_i\sum\limits_j s_{ij}\|\textbf{x}_i - \textbf{x}_j\|_2^2 \\
			%			&=  \sum\limits_i\sum\limits_j d_{ij}<\textbf{f}_i, \textbf{f}_j>\\
			%			&= \textrm{tr}(\textbf{F}^\textrm{T}\textbf{D}\textbf{F}) \\
			%			&= \textrm{tr}(\textbf{Y}^\textrm{T}\textbf{D}\textbf{Y}\textbf{P}^{-1})
		\end{aligned}
	\end{equation}
	Now Lemma~\ref{lemma1} is proven.
\end{proof}

	Our approach constructs a manifold structure \textbf{S} from label \textbf{Y}, thereby ensuring label consistency for samples on the same manifold. This also eliminates the need to estimate the centroid matrix.
	By Lemma~\ref{lemma1}, Eq.~(\ref{rsfkm_euc}) can be equivalent to Eq.~(\ref{fkmwc1}). The objective function of our FKMWC is delineated below:
	\begin{equation}\label{fkmwc1}
		\begin{aligned}
			\mathop{\textrm{min}}\limits_{\textbf{Y}} \sum\limits_i\sum\limits_j s_{ij}\|\textbf{x}_i - \textbf{x}_j\|_2^2 + \lambda \left\|\textbf{Y}\right\|_F^2 \\
			s.t. \quad  s_{jj} = \sum_{i}{y_{ij}}, \textbf{Y}\textbf{1}=\textbf{1}, \textbf{Y}\ge\textbf{0}
		\end{aligned}
	\end{equation}
	
	%To address these issues, we propose a modified approach: Fuzzy \textit{K}-Means clustering without cluster centroids (\textbf{FKMWC}). FKMWC obviates the need for selecting and updating cluster centroids. The objective function of FKMWC is delineated below:
	
	In order to facilitate the solution, we convert Eq.~(\ref{fkmwc1}) into the form of Eq.~(\ref{fkmwc}) for solving
	
	\begin{equation}\label{fkmwc}
		\begin{aligned}
			& \sum\limits_i\sum\limits_j s_{ij}\|\textbf{x}_i - \textbf{x}_j\|_2^2 + \lambda \left\|\textbf{Y}\right\|_F^2\\
			&\quad\quad=  \sum\limits_i\sum\limits_j d_{ij}<\textbf{f}_i, \textbf{f}_j>+ \lambda \left\|\textbf{Y}\right\|_F^2\\
			&\quad\quad= \textrm{tr}(\textbf{F}^\textrm{T}\textbf{D}\textbf{F}) + \lambda \left\|\textbf{Y}\right\|_F^2\\
			&\quad\quad= \textrm{tr}(\textbf{Y}^\textrm{T}\textbf{D}\textbf{Y}\textbf{P}^{-1})+ \lambda \left\|\textbf{Y}\right\|_F^2 \\
			&s.t. \quad  p_{jj} = \sum_{i}{y_{ij}}, \textbf{Y}\textbf{1}=\textbf{1}, \textbf{Y}\ge\textbf{0}
		\end{aligned}
	\end{equation}

where $\mathbf{P}$ is a diagonal matrix and $\mathbf{D}$ is the distance matrix. If the matrix $\mathbf{D}$ in Eq.~(\ref{fkmwc}) represents the squared Euclidean distance, that is, $d_{ij} = \left\|\mathbf{x}_i - \mathbf{x}_j\right\|_2^2$, then Eq.~(\ref{fkmwc}) becomes equivalent to Eq.~(\ref{rsfkm_euc}) as per Lemma~\ref{lemma1}. This implies that the objective of model~(\ref{fkmwc}) aligns with that of the Fuzzy $K$-Means Clustering.

In our algorithm, clusters are entirely eliminated, and the membership matrix $\mathbf{Y}$ is derived solely from the distance matrix $\mathbf{D}$. This strategy significantly enhances the robustness of the clustering performance and offers increased flexibility in the measurement of distances.

The FKMWC proposed by us can serve as a universal framework for Fuzzy \textit{K}-Means clustering. Due to its decoupling of distance computation and optimization solving for membership matrices, FKMWC can conveniently adapt to different characteristics of datasets when applied, allowing for flexible selection of various distance metrics. This enables FKMWC to be easily extended to other improved Fuzzy \textit{K}-Means clustering algorithms. For instance, for linearly inseparable datasets, FKMWC can be extended to kernel methods by computing kernel distances. Additionally, FKMWC is also applicable when only an adjacency graph matrix is provided as input, requiring only the conversion of the adjacency graph matrix into a distance matrix.

In particular, when $\lambda = 0$, the objective function~(\ref{fkmwc}) can only be minimized if each sample is assigned to the nearest cluster center. In this case, FKMWC is equivalent to the classical \textit{K}-Means clustering.

\subsection{Distance matrix}

The FKMWC algorithm inputs a fixed global distance matrix and directly optimizes to obtain the fuzzy membership matrix, thus allowing more flexibility in measuring sample distances $\textbf{D}$. Below are some examples:

\subsubsection{\textbf{Squared Euclidean distance}}

The squared Euclidean distance measures the absolute distance between points in a Euclidean space. The distance between the $i$-th and $j$-th samples can be expressed as follows:
\begin{equation}\label{EucD}
	\begin{aligned}
		d_{euc(ij)} = \left\|\textbf{x}_i -\textbf{x}_j\right\|_2^2
	\end{aligned}
\end{equation}

\subsubsection{\textbf{K-Nearest Neighbor Distance}}

The \textit{K-nearest neighbor distance} refers to the process of computing the distances between a given query point and all other points in the dataset, and then selecting the \textit{K} nearest points.
\begin{equation}\label{knnEucD}
	\begin{aligned}
		d_{knn(ij)} = \begin{cases}
			\|\textbf{x}_i-\textbf{x}_j\|_2^2, &\textbf{x}_i\leftrightarrow \textbf{x}_j\\
			\sigma,&\text{otherwise}
		\end{cases}
	\end{aligned}
\end{equation}
where $\textbf{x}_i\leftrightarrow \textbf{x}_j$ means that $\textbf{x}_i$ is in the set of $k$ nearest neighbors of $\textbf{x}_j$ and $\textbf{x}_j$ is in the set of $k$ nearest neighbors of $\textbf{x}_i$, $\sigma$ can be set as the distance between the two farthest samples in the dataset.

\subsubsection{\textbf{Butterworth Distance}}

Given an adjacency matrix $\mathbf{S}$, the distance matrix $\mathbf{D}$ can be computed utilizing a Butterworth filter as proposed in \cite{LuGWYX23}. The process is outlined as follows:
\begin{equation}\label{adj_D}
	\begin{aligned}
		d_{bw(ij)}=\sqrt{\frac1{1+(\frac{s_{ij}}\Omega)^4}}
	\end{aligned}
\end{equation}
where $\Omega$ is a hyperparameter

\subsubsection{\textbf{Kernel Distance}}
Kernel $\textit{K}$-Means clustering is a clustering analysis algorithm that leverages kernel methods. The core concept involves mapping data, denoted as $\mathbf{x}$, into a higher-dimensional space represented by $\phi(\mathbf{x})$. The algorithm computes the similarity between data points using kernel functions. For instance, the Gaussian radial basis function is defined as:
\[
K(\mathbf{x}, \mathbf{y}) = \exp\left(-\frac{\|\mathbf{x}-\mathbf{y}\|_2^2}{2\sigma^2}\right),
\]
where $\|\mathbf{x}-\mathbf{y}\|_2$ denotes the Euclidean distance between $\mathbf{x}$ and $\mathbf{y}$, and $\sigma$ is a scaling parameter. Our method can be extended to other kernel methods by incorporating kernel distances as inputs. These kernel distances are computed as follows:
\begin{equation}\label{kernel_D}
	\begin{aligned}
		d_{ker(ij)} &= \|\phi(\textbf{x}_i)-\phi(\textbf{x}_j)\|_2^2 \\
		&= K(\textbf{x}_i,\textbf{x}_i) + K(\textbf{x}_j,\textbf{x}_j) - 2K(\textbf{x}_i,\textbf{x}_j)
	\end{aligned}
\end{equation}

\subsection{Optimization Algorithm}

We use gradient descent to solve Eq.~(\ref{fkmwc}), if we have
\begin{equation}\label{solY_J}
	\begin{aligned}
		J(\textbf{Y}) = \textrm{tr}(\textbf{Y}^\textrm{T}\textbf{D}\textbf{Y}\textbf{P}^{-1}) + \lambda \left\|\textbf{Y}\right\|_F^2
	\end{aligned}
\end{equation}
then
\begin{equation}\label{grad_J}
	\begin{aligned}
		\frac{\partial J}{\partial \textbf{Y}} = (\textbf{D} + \textbf{D}^\textrm{T})\textbf{YP}^{-1} + 2\lambda\textbf{Y} + \frac{\partial (\textbf{a}^\textrm{T}\hat{\textbf{p}})}{\partial \textbf{Y}}
	\end{aligned}
\end{equation}
where $a_i = (\textbf{Y}^\textrm{T}\textbf{D}\textbf{Y})_{ii}$, and $\hat{p} _i = p_{ii}^{-1}$,
for $(i,j)$-th entry of \textbf{Y}, since $\frac{\partial p_{jj}}{\partial y_{ij}}=1,$ it is clear that
\begin{equation}\label{grad_ap}
	\begin{aligned}
		\frac{\partial (\textbf{a}^\textrm{T}\hat{\textbf{p}})}{\partial y_{ij}} =  \frac{\partial}{\partial y_{ij}} \left(\sum\limits_{j = 1}^{K}{a_j p_{jj}^{-1}}\right) = -a_j p_{jj}^{-2}
	\end{aligned}
\end{equation}

Let $\textbf{G} = (\textbf{D} + \textbf{D}^\textrm{T})\textbf{YP}^{-1} + 2\lambda\textbf{Y}$, for any $i$, we have
\begin{equation}\label{grad_J}
	\begin{aligned}
		\frac{\partial J}{\partial y_{ij}} = g_{ij} - a_j p_{jj}^{-2}
	\end{aligned}
\end{equation}

Then minimizing $J$ is equivalent to iteratively solving the following:
\begin{equation}\label{grad_H}
	\begin{aligned}
		y_{ij} \leftarrow y_{ij}-\eta\nabla J
	\end{aligned}
\end{equation}
taking $\eta = \frac{y_{ij}}{g_{ij} + \sqrt{g_{ij}a_j p_{jj}^{-2}}}$, and then

\begin{equation}\label{iterY}
	\begin{aligned}
		y_{ij} \leftarrow y_{ij}\sqrt{\frac{a_j p_{jj}^{-2} }{g_{ij}}}
	\end{aligned}
\end{equation}

%The code version:
%\begin{equation}\label{iterY}
%	\begin{aligned}
	%		Y_{ij} \leftarrow Y_{ij}\sqrt{\frac{a_j %p_{jj}^{-2} + \sum_j G_{ij} / K}{G_{ij} + %\sum_j{(a_j p_{jj}^{-2})} / K}}
	%\end{aligned}
	%\end{equation}

	The Algorithm~\ref{A1} lists the pseudo code of solving (\ref{fkmwc}).
	
	\begin{algorithm}[!t]
		\caption{Solving the Model (\ref{fkmwc})}
		% \LinesNumbered
		\begin{algorithmic}[1]\label{A1}
			\REQUIRE The distance matrix $\mathbf{D}$.\\
			\ENSURE The membership matrix $\mathbf{Y}$.\\
			\STATE \textbf{Initialize}: $\mathbf{Y}$, $\lambda$.\\	
			\WHILE{\emph{not converge}}
			\STATE Calculate $a_i = (\textbf{Y}^\textrm{T}\textbf{D}\textbf{Y})_{ii}$
			\STATE Calculate $p_{jj} = \sum_{i}{y_{ij}}$
			\STATE Calculate $\textbf{G} = (\textbf{D} + \textbf{D}^\textrm{T})\textbf{YP}^{-1} + 2\lambda\textbf{Y}$
			\STATE Update $\textbf{Y}$ via \eqref{iterY}
			\STATE Normalize $\textbf{Y}$ by $y_{ij} = \frac{y_{ij}}{\sum\limits_j{y_{ij}}}$
			\ENDWHILE
			\STATE \textbf{return}: The membership matrix $\mathbf{Y}$
		\end{algorithmic}
	\end{algorithm}

	\subsection{Computational Complexity Analysis}
	The computational complexity of the proposed method involves computations across four variables within a single iterative step. As outlined in Algorithm~\ref{A1}, the calculation of $\textbf{a}$ requires $\bm{\mathcal{O}}(K(N^2+N))$ operations. Similarly, the computation of $\textbf{P}$ demands $\bm{\mathcal{O}}(KN)$ operations, while determining $\textbf{G}$ necessitates $\bm{\mathcal{O}}(KN^2)$ operations. Additionally, the update of $\textbf{Y}$ is achieved with $\bm{\mathcal{O}}(KN)$ operations. Consequently, the predominant computational complexity of our approach is represented as $\bm{\mathcal{O}}(TKN^2)$, where $K$ denotes the number of clusters, $N$ symbolizes the number of samples, and $T$ signifies the number of iterations.
	
	\section{Experiments}
	
	The experiments are conducted on a Windows 11 desktop computer equipped with a Intel(R) Core(TM) i5-13600KF 3.50GHz CPU and 32 GB RAM. The code is executed in MATLAB R2023a (64-bit).
	
						\subsection{Benchmark Datasets}
	The performance of our method is evaluated on seven benchmark datasets, and the information of size is shown in Table~\ref{datasets_info}.
		\begin{enumerate}
	\item\textbf{AR}~\cite{AleixAR1998} includes 120 kinds of face with 3120 images.
	\item \textbf{JAFFE}~\cite{670949} consists of 213 images of different facial expressions from 10 different Japanese female subjects.
	\item\textbf{MSRC\_V2}~\cite{WinnJ05} which includes 7 kinds of objects with 210 images. We selected the 576-D HOG feature as the single view dataset.
	\item\textbf{ORL}~\cite{Cai2010UnsupervisedFS} includes 400 pictures of face from 40 people.
	\item\textbf{UMIST}~\cite{6565365} consists of 564 images of face from 20 individuals
	\item\textbf{USPS}~\cite{291440} consists of 3000 handwritten digital images for each of the 10 digits from 0 to 9.
\item	\textbf{Yaleface}~\cite{598228} includes 165 grayscale images in GIF format of 15 people.
\end{enumerate}
	\begin{table}[ht]
		\centering
		\caption{The information of the benchmark datasets, including }\label{datasets_info}
		\begin{center}
			 \resizebox{0.8\columnwidth}{!}{
				\begin{tabular}{l c c c }
					\toprule
					% Datasets &\multicolumn{3}{c|}{MSRC} &\multicolumn{3}{c|}{HW} &\multicolumn{3}{c|}{ORL} &\multicolumn{3}{c}{Mnist4}\\
					% \midrule
					Datasets & Sample & dimension & clusters \\
					\midrule
					AR			& 3120	& 2000	& 120	\\
					JAFFE		& 213	& 676	& 10	\\
					MSRC\_V2	& 210	& 576	& 7	    \\
					ORL			& 400	& 1024	& 40	\\
					UMIST		& 575	& 1024	& 20	\\
					USPS		& 3000	& 256	& 10	\\
					Yaleface	& 165	& 1024	& 15	\\
					\bottomrule
				\end{tabular}}
			\end{center}
		\end{table}

		\begin{table*}[!t]
			\centering
			\caption{ACC results on 7 benchmark datasets from 7 comparative algorithms and FKMWC with different distance, with the highest value represented in bold.}\label{res_acc}
			\begin{center}
				 \resizebox{2.05\columnwidth}{!}{
					\begin{tabular}{l c c c c c c c | c c c c}
						\toprule
						% Datasets &\multicolumn{3}{c|}{MSRC} &\multicolumn{3}{c|}{HW} &\multicolumn{3}{c|}{ORL} &\multicolumn{3}{c}{Mnist4}\\
						% \midrule
						Datasets & \textit{K}-Means++ & Ksum & Ksumx &  RKM & CDKM & MSFCM & ULGE  & Ours\_bw & Ours\_d & Ours\_ker & Ours\_kd\\
						\midrule
						AR			& 0.2514	& 0.2970	& 0.2454	& 0.2641	& 0.2653	& 0.2747	& 0.3442 & \textbf{0.3917} & 0.2843 &0.2891	& 0.3728 \\
						JAFFE		& 0.7085	& 0.8789	& 0.8930	& 0.8310	& 0.7108	& 0.8357	& 0.9249 & 0.9155 & 0.8404 & 0.9108 & \textbf{0.9671} \\
						MSRC\_V2	& 0.6052	& 0.7524	& 0.6852	& 0.6286	& 0.6657	& 0.7381	& 0.7429 & 0.7714 & 0.7667 & 0.7619 & \textbf{0.8095} \\
						ORL			& 0.5198	& 0.6337	& 0.5877	& 0.5000	& 0.5507	& 0.5725	& 0.6200 & 0.6625 & 0.6075 & 0.6250 & \textbf{0.6875} \\
						UMIST		& 0.4339	& 0.4209	& 0.4296	& 0.4209	& 0.4210	& 0.4591	& 0.5043 & 0.4974 & 0.4713 & 0.5113 & \textbf{0.5252} \\
						USPS		& 0.6491	& 0.7664	& 0.7240	& 0.6673	& 0.6416	& 0.6577	& 0.7530 & \textbf{0.7923} & 0.7013 & 0.7110& 0.7553 \\
						Yaleface	& 0.3812	& 0.4339	& 0.4418	& 0.4485	& 0.3964	& 0.4545	& 0.5212 & 0.4909 & 0.4788 & 0.4909 & \textbf{0.5273} \\
						\bottomrule
					\end{tabular}}
				\end{center}
			\end{table*}
			
			%by Eq.~(\ref{adj_D}) (Ours\_bw) ,squared Euclid distance (Ours\_d) and K nearest neighbors distance (Ours\_kd)
			\begin{table*}[!t]
				\centering
				\caption{NMI results on 7 benchmark datasets from 7 comparative algorithms and FKMWC with different distance, with the highest value represented in bold.}\label{res_nmi}
				\begin{center}
					 \resizebox{2.05\columnwidth}{!}{
						\begin{tabular}{l c c c c c c c | c c c c}
							\toprule
							% Datasets &\multicolumn{3}{c|}{MSRC} &\multicolumn{3}{c|}{HW} &\multicolumn{3}{c|}{ORL} &\multicolumn{3}{c}{Mnist4}\\
							% \midrule
							Datasets & \textit{K}-Means++ & Ksum & Ksumx &  RKM & CDKM & MSFCM & ULGE &Ours\_bw& Ours\_d & Ours\_ker & Ours\_kd \\
							\midrule
							AR		& 0.5574	& 0.5963	& 0.5676	& 0.5752	& 0.5700	& 0.5811	& 0.6090 & \textbf{0.6500} & 0.5722 & 0.5766 & 0.6268 \\
							JAFFE	& 0.8010	& 0.8764	& 0.9013	& 0.8159	& 0.7981	& 0.8806	& 0.9005 & 0.9225 & 0.8545 & 0.9175 & \textbf{0.9623} \\
							MSRC\_V2& 0.5280	& 0.6111	& 0.5753	& 0.5612	& 0.5693	& 0.6010	& 0.5870 & 0.6265 & 0.6475 & 0.6044 & \textbf{0.6712} \\
							ORL		& 0.7234	& 0.7940	& 0.7693	& 0.7143	& 0.7529	& 0.7697	& 0.8052 & 0.8018 & 0.7813 & 0.7850 & \textbf{0.8053} \\
							UMIST	& 0.6410	& 0.6190	& 0.6377	& 0.5963	& 0.6404	& 0.6469	& 0.6420 & 0.6719 & 0.6503 & 0.6759 & \textbf{0.6791} \\
							USPS	& 0.6147	& 0.6615	& 0.6078	& 0.5865	& 0.6059	& 0.5989	& 0.6306 & \textbf{0.6768} & 0.5833 & 0.6192 & 0.6440 \\
							Yaleface& 0.4389	& 0.4975	& 0.5018	& 0.5099	& 0.4779	& 0.5064	& 0.5356 & 0.5382 & \textbf{0.5461} & 0.5439 & 0.5330 \\
							\bottomrule
						\end{tabular}}
					\end{center}
				\end{table*}
				
				\begin{table*}[!t]
					\centering
					\caption{Purity results on 7 benchmark datasets from 7 comparative algorithms and FKMWC with different distance, with the highest value represented in bold.}\label{res_pur}
					\begin{center}
						 \resizebox{2.05\columnwidth}{!}{
							\begin{tabular}{l c c c c c c c | c c c c}
								\toprule
								% Datasets &\multicolumn{3}{c|}{MSRC} &\multicolumn{3}{c|}{HW} &\multicolumn{3}{c|}{ORL} &\multicolumn{3}{c}{Mnist4}\\
								% \midrule
								Datasets & \textit{K}-Means++ & Ksum & Ksumx &  RKM & CDKM & MSFCM & ULGE & Ours\_bw & Ours\_d & Ours\_ker & Ours\_kd\\
								\midrule
								AR		& 0.2749	& 0.3686	& 0.3236	& 0.3215	& 0.2862	& 0.2942	& 0.3676 & \textbf{0.4071} & 0.3006 & 0.3080 & 0.3917 \\
								JAFFE	& 0.7455	& 0.8789	& 0.8977	& 0.8310	& 0.7441	& 0.8404	& 0.9249 & 0.9155 & 0.8404 & 0.9108 & \textbf{0.9671} \\
								MSRC\_V2& 0.6276	& 0.7524	& 0.6910	& 0.6333	& 0.6795	& 0.7381	& 0.7429 & 0.7714 & 0.7667 & 0.7619 & \textbf{0.8095} \\
								ORL		& 0.5705	& 0.6562	& 0.6060	& 0.5200	& 0.6090	& 0.6250	& 0.6700 & 0.6900 & 0.6425 & 0.6675 & \textbf{0.7100} \\
								UMIST	& 0.5110	& 0.4553	& 0.4715	& 0.4400	& 0.5043	& 0.5113	& 0.5322 & 0.5443 & 0.5270 & 0.5635 & \textbf{0.5809} \\
								USPS	& 0.6803	& 0.7677	& 0.7240	& 0.6827	& 0.6746	& 0.6717	& 0.7530 & \textbf{0.7923} & 0.7013 & 0.7110 & 0.7553 \\
								Yaleface& 0.4030	& 0.4733	& 0.4915	& 0.4848	& 0.4188	& 0.4606	& 0.5212 & 0.5030 & 0.4788 & 0.4909 & \textbf{0.5333} \\
								
								\bottomrule
							\end{tabular}}
						\end{center}
					\end{table*}
						\begin{figure*}[!h]
						% \subfigure[AR]{
							% 	\includegraphics[width=0.32\linewidth]{figures/L2_AR_lambda.pdf}
							% }
						\subfigure[JAFFE (Ours\_d)]{
							\includegraphics[width=0.32\linewidth]{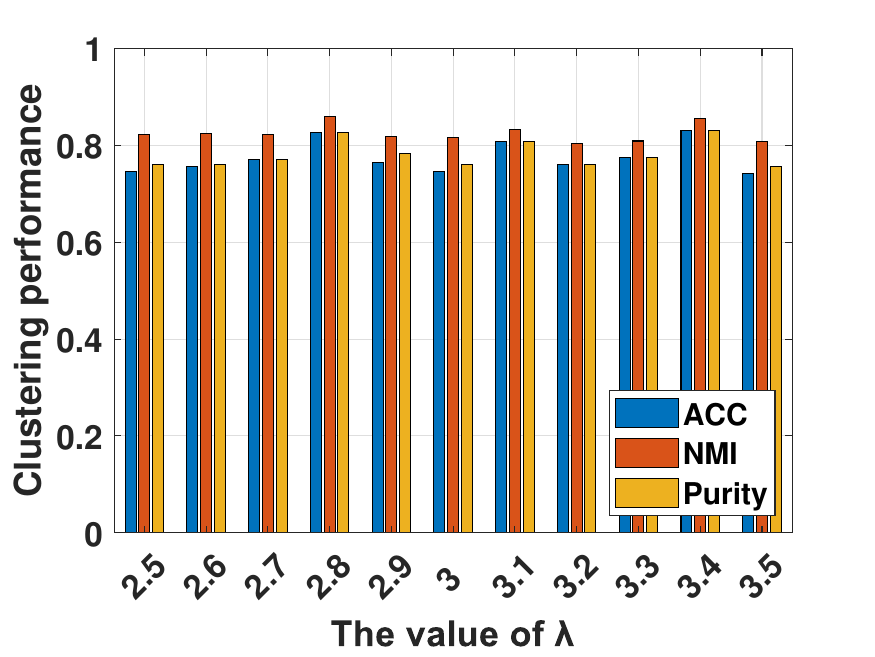}
						}
						\subfigure[ORL (Ours\_d)]{
							\includegraphics[width=0.32\linewidth]{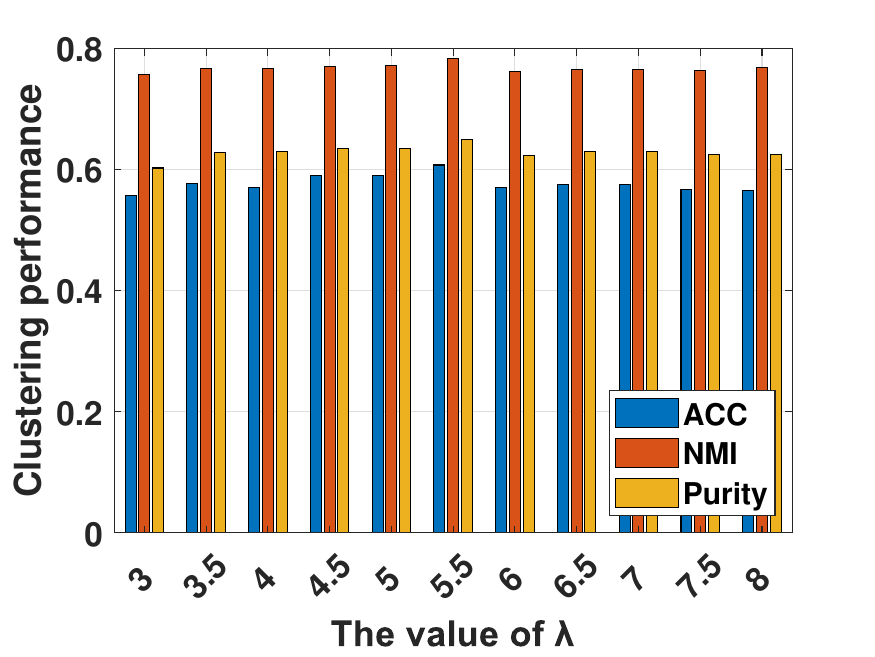}
						}
						\subfigure[MSRC\_V2 (Ours\_d)]{
							\includegraphics[width=0.32\linewidth]{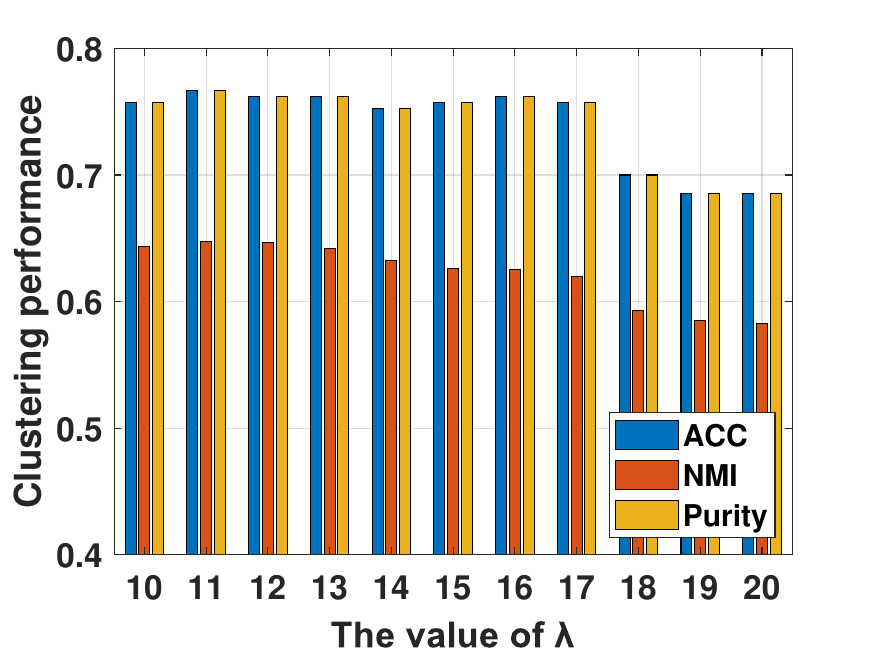}
						}
						\subfigure[UMIST (Ours\_d)]{
							\includegraphics[width=0.32\linewidth]{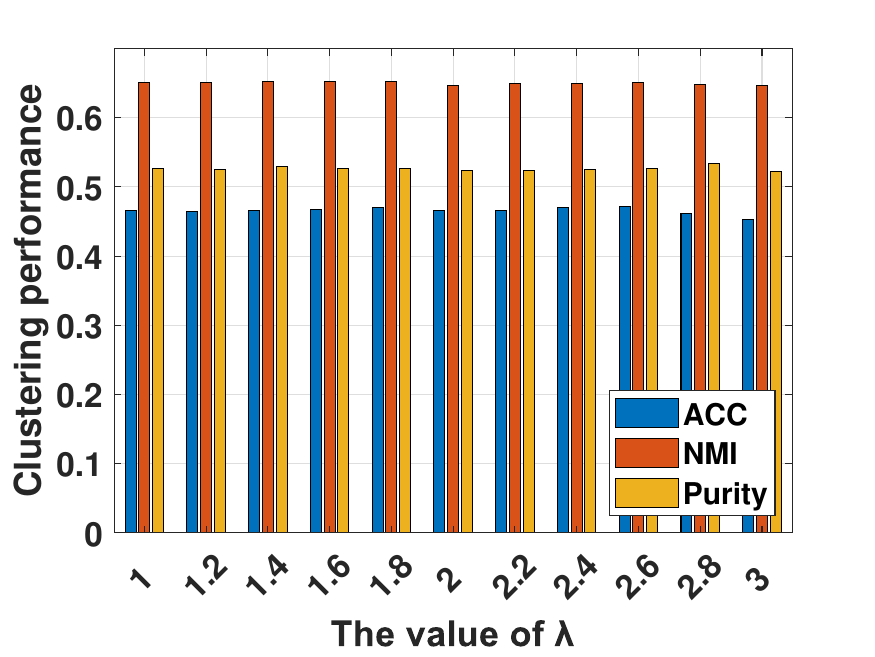}
						}
						\subfigure[USPS (Ours\_d)]{
							\includegraphics[width=0.32\linewidth]{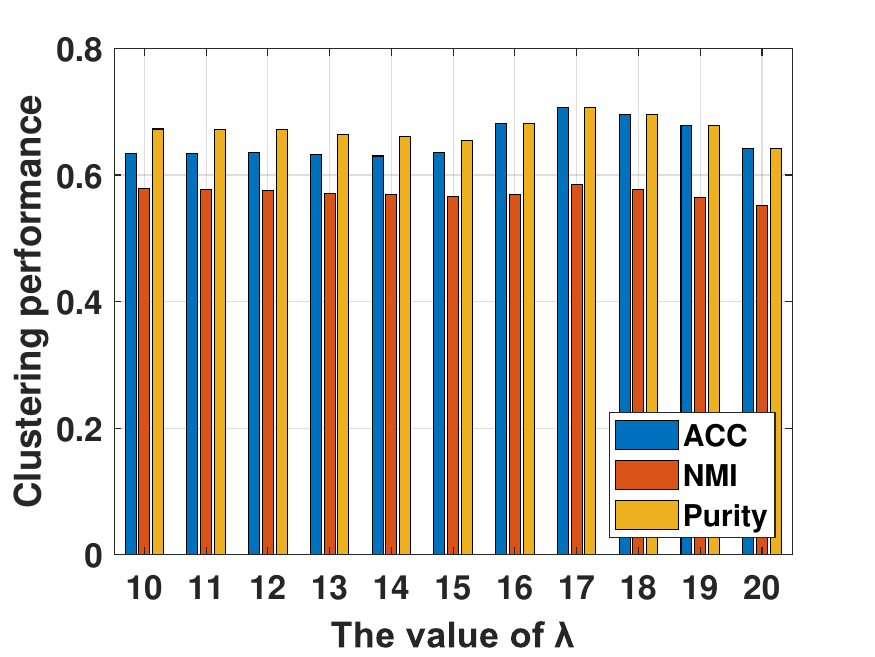}
						}
						\subfigure[Yaleface (Ours\_d)]{
							\includegraphics[width=0.32\linewidth]{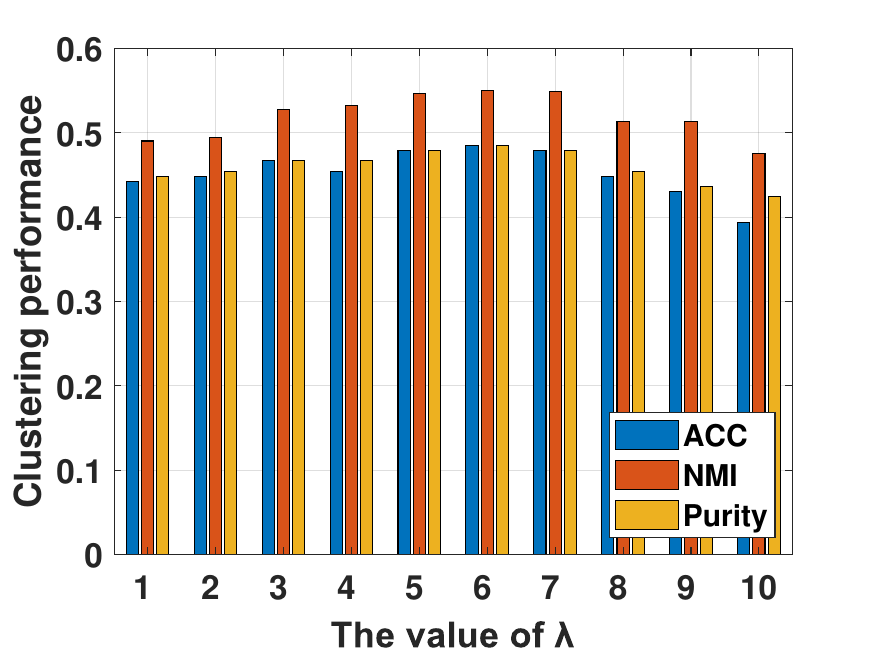}
						}
						\subfigure[JAFFE (Ours\_kd)]{
							\includegraphics[width=0.32\linewidth]{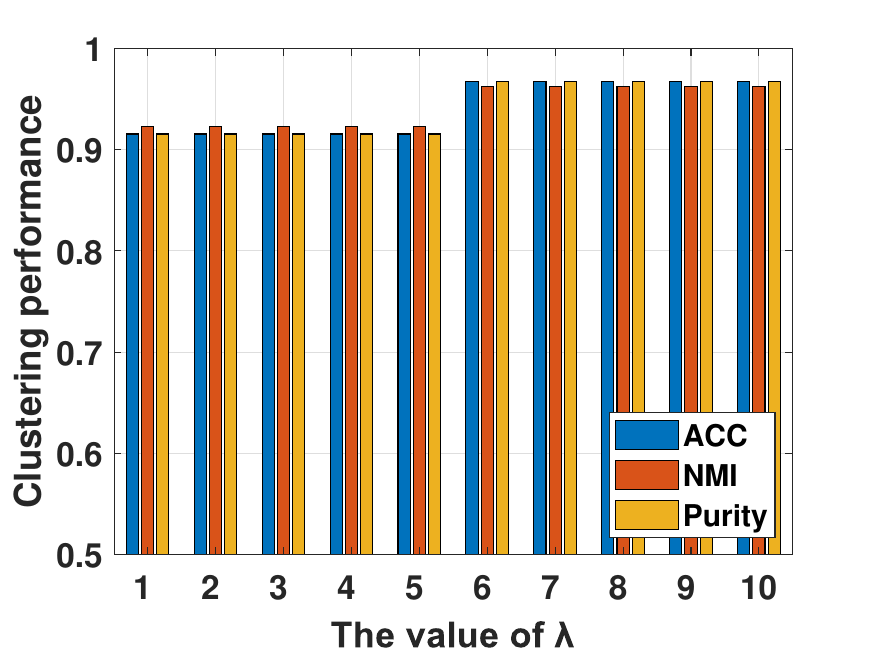}
						}
						\subfigure[ORL (Ours\_kd)]{
							\includegraphics[width=0.32\linewidth]{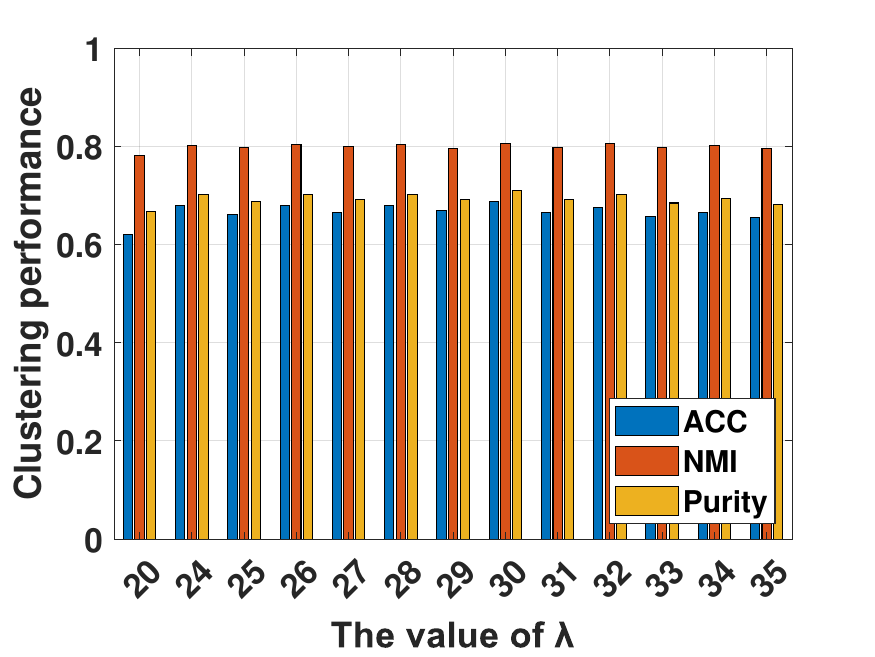}
						}
						\subfigure[MSRC\_V2 (Ours\_kd)]{
							\includegraphics[width=0.32\linewidth]{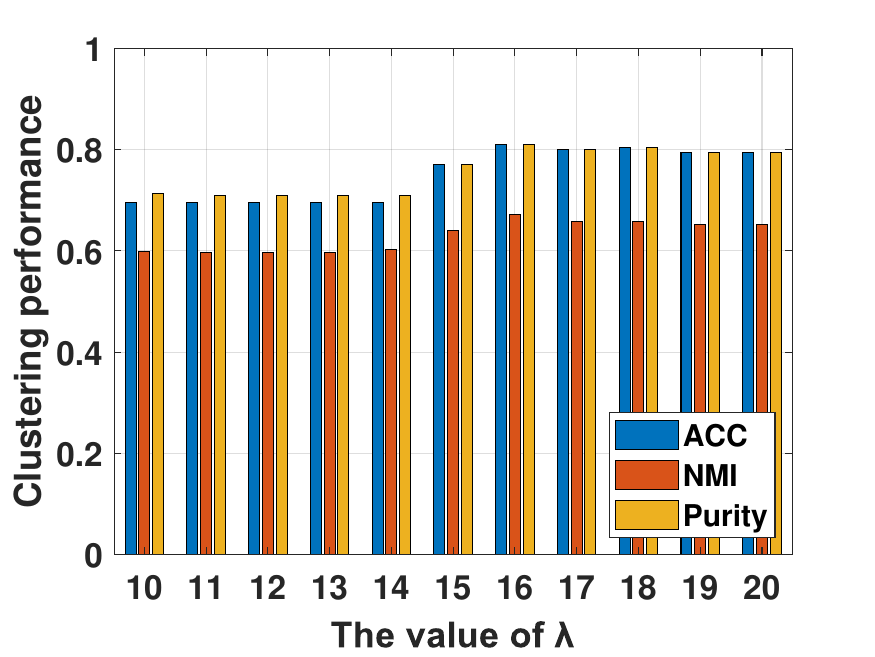}
						}
						\subfigure[UMIST (Ours\_kd)]{
							\includegraphics[width=0.32\linewidth]{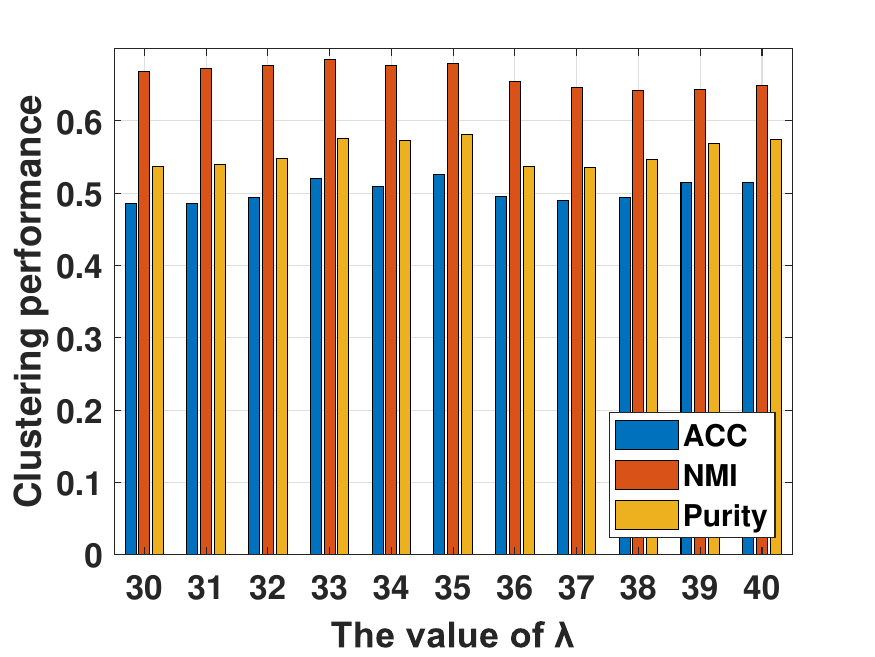}
						}
						\subfigure[USPS (Ours\_kd)]{
							\includegraphics[width=0.32\linewidth]{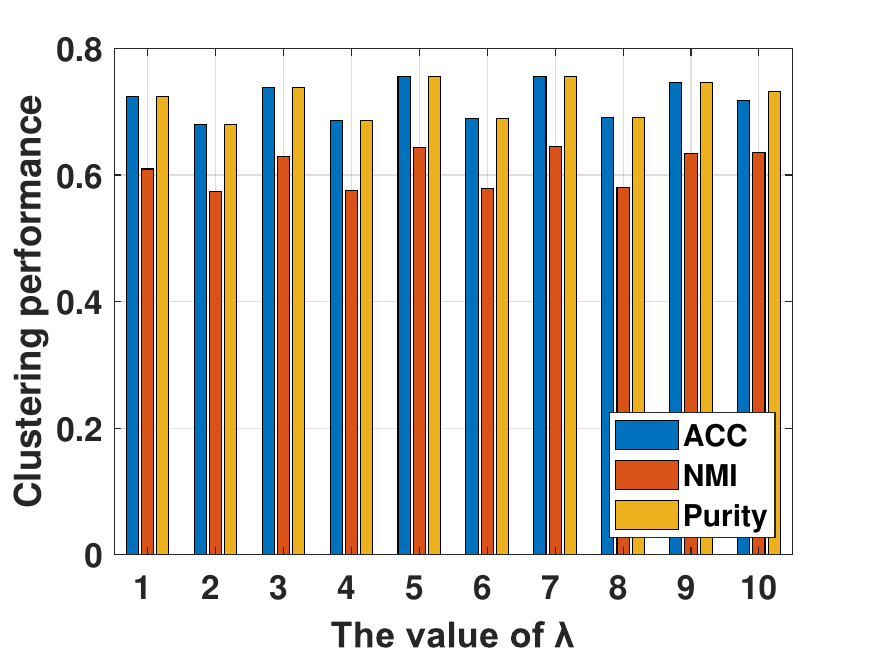}
						}
						\subfigure[Yaleface (Ours\_kd)]{
							\includegraphics[width=0.32\linewidth]{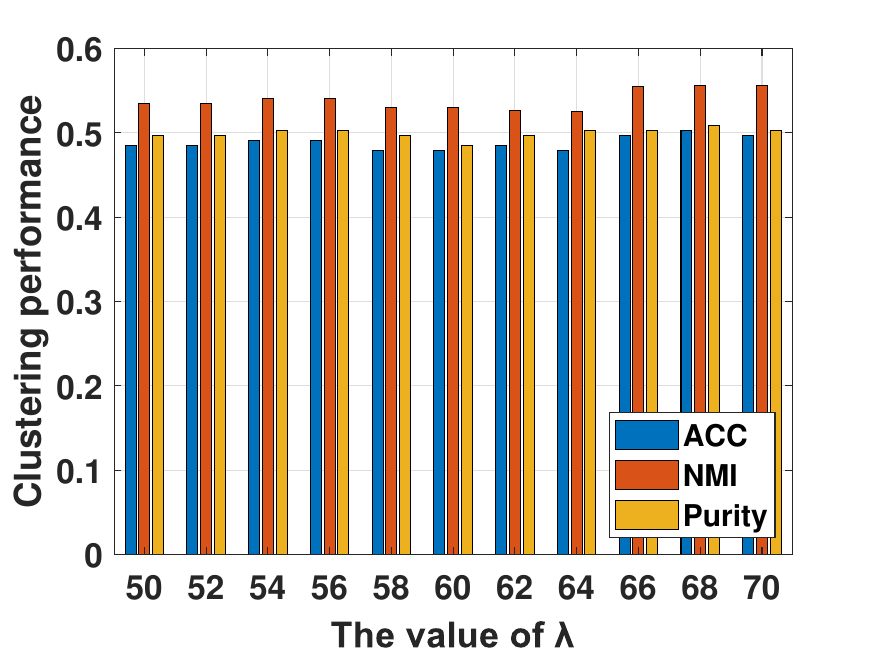}
						}
						\caption{Performance of the FKMWC vs. $\lambda$ on 6 benchmark datasets with squared Euclidean distance (Ours\_d) and K-nearest neighbor distance (Ours\_kd).}
						\label{result_lambda}
					\end{figure*}

					\begin{figure*}
						% \subfigure[AR]{
							% 	\includegraphics[width=0.32\linewidth]{figures/L2_AR_lambda.pdf}
							% }
						\subfigure[JAFFE]{
							\includegraphics[width=0.32\linewidth]{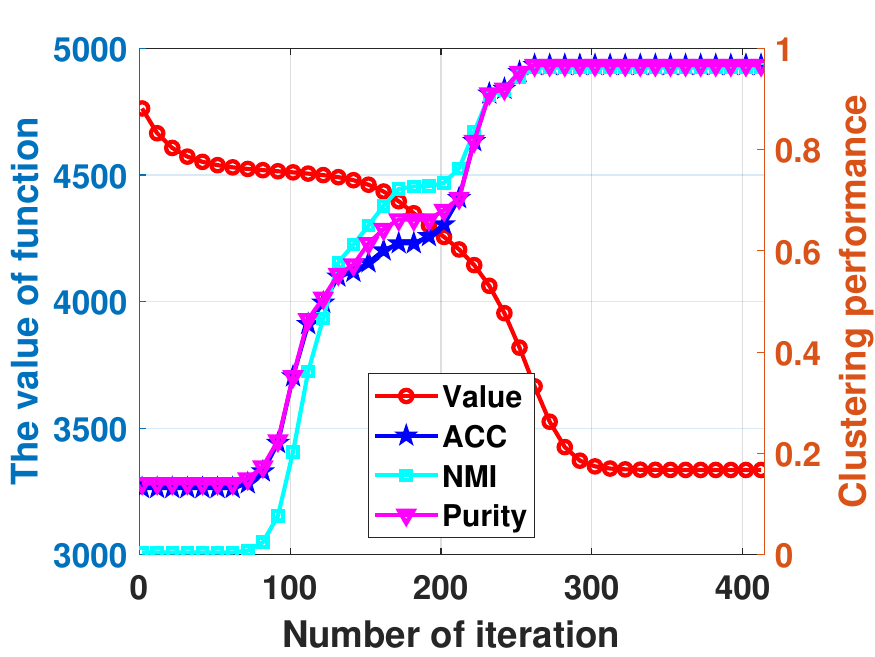}
						}
						\subfigure[ORL]{
							\includegraphics[width=0.32\linewidth]{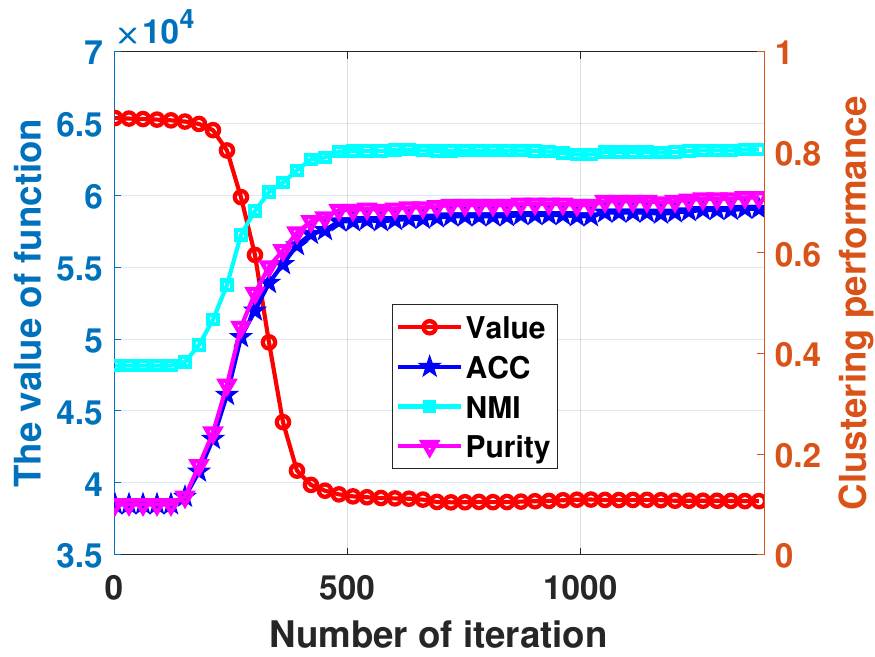}
						}
						\subfigure[MSRC\_V2]{
							\includegraphics[width=0.32\linewidth]{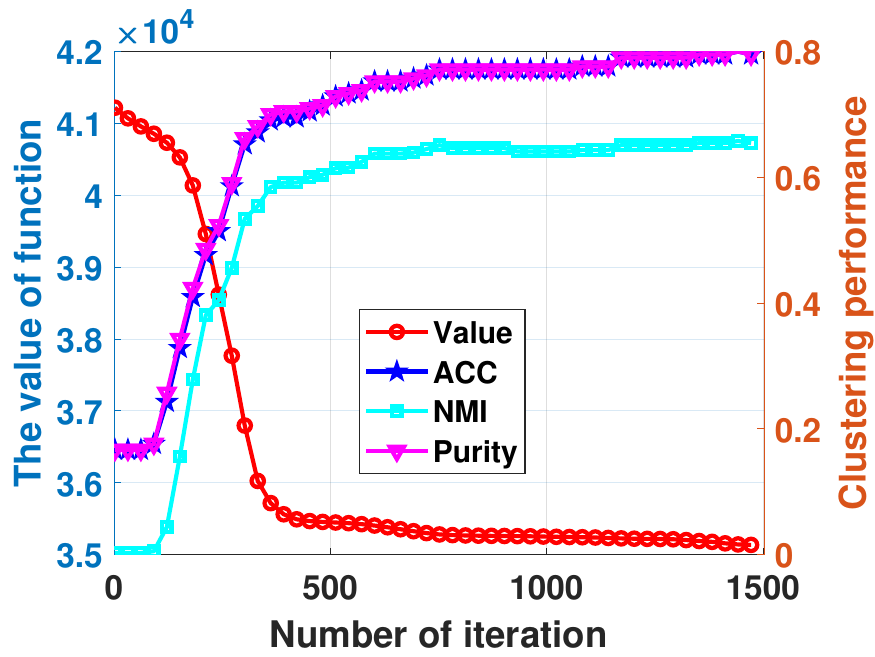}
						}
						\subfigure[UMIST]{
							\includegraphics[width=0.32\linewidth]{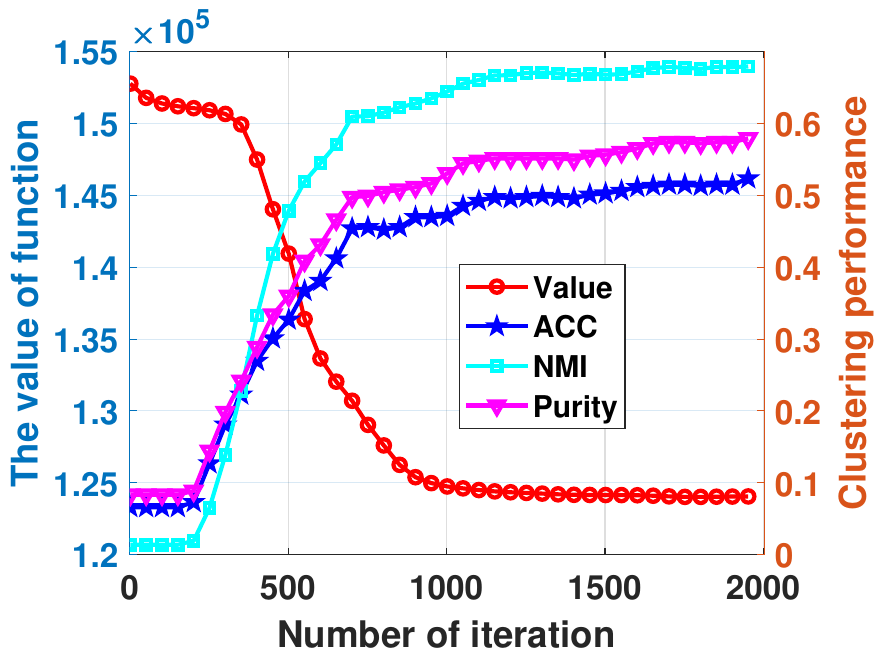}
						}
						\subfigure[USPS]{
							\includegraphics[width=0.32\linewidth]{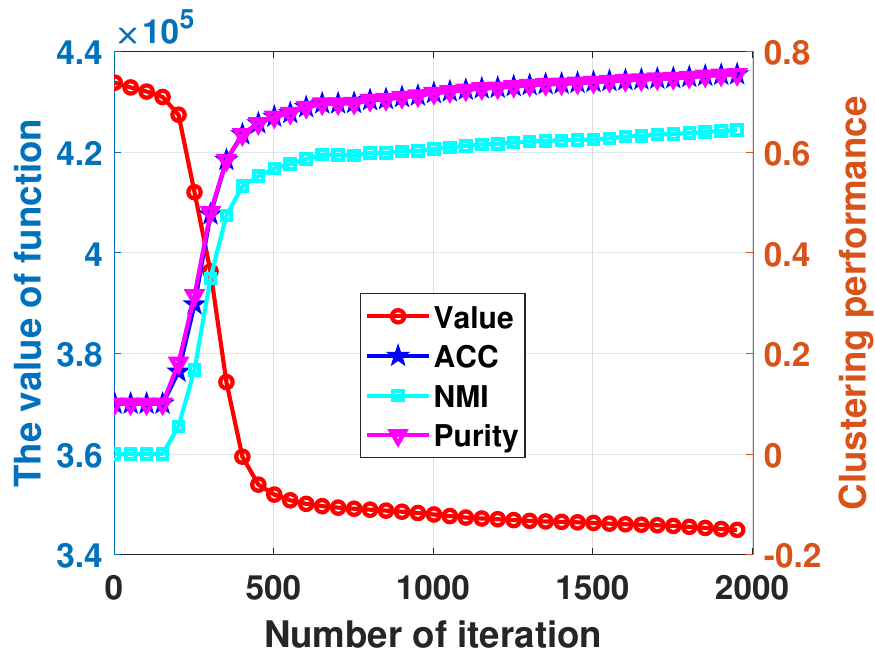}
						}
						\subfigure[Yaleface]{
							\includegraphics[width=0.32\linewidth]{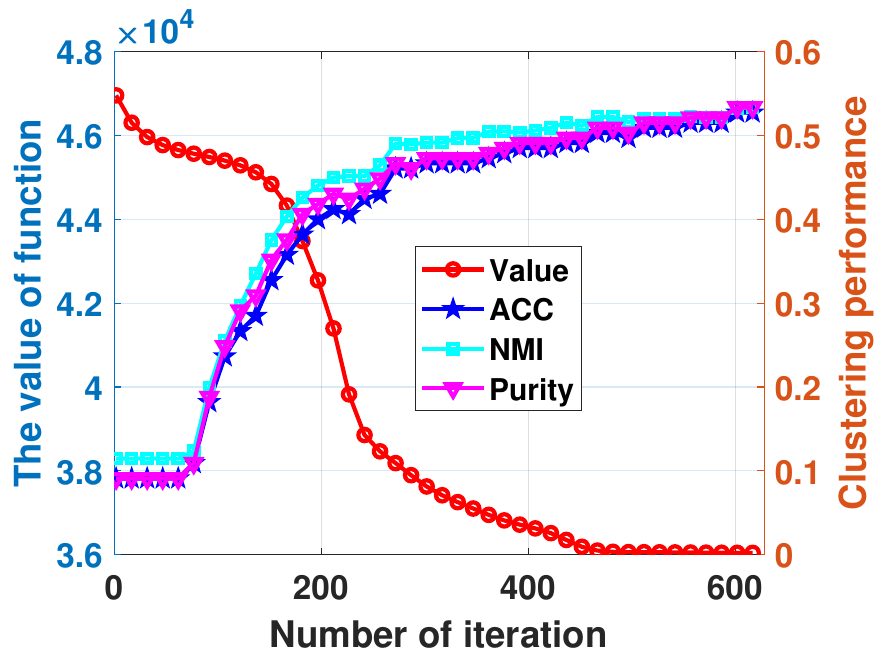}
						}
						\caption{The value of objective function~(\ref{fkmwc}) and clustering performance with iterations on 6 benchmark datasets.}
						\label{result_iter}
					\end{figure*}

					\subsection{Comparison Algorithms}
						\begin{enumerate}
					\item\textbf{\emph{K-Means++}}~\cite{ArthurV07}:The advantages of careful seeding for \textit{K}-Means.
					\item\textbf{Ksum}~\cite{Ksum}: This algorithm integrates spectral clustering and \emph{K}-Means into a unified framework.
					\item\textbf{Ksum-x}~\cite{Ksum}: Distinct from Ksum, Ksum-x inputs a data matrix instead of a K-NN graph.
					\item\textbf{RKM}~\cite{ijcai2019p414}: This algorithm minimizes the squared distance from the samples to the clustering centers and incorporates a regularization term for balanced clustering.
					\item\textbf{CDKM}~\cite{9444882}: Demonstrates that the underlying model of \emph{K}-Means can be transformed into a matrix trace maximization problem, proposing an efficient coordinate descent solution.
					\item\textbf{MSFCM}~\cite{ZhouLZP21}A new membership scaling Fuzzy C-Means clustering algorithm.
					\item\textbf{ULGE}~\cite{NieZL17} Unsupervised Large Graph Embedding.
						\end{enumerate}
					\subsection{Metrics}
					The performance of our method is evaluated by the following three metrics.
					\begin{enumerate}
						\item \emph{\textbf{Accuracy (ACC)}}: It represents the clustering accuracy, defined as:
						\begin{equation}\label{eq:acc}
							ACC = \frac{\sum_{i=1}^{n} \delta (a_i, \text{map}(b_i))}{n}
						\end{equation}
						where $a_i$ is the true label, $b_i$ is the predicted label, $\text{map}(\cdot)$ maps the predicted label to the true label, and $\delta (x,y)$ is an indicator function:
						\begin{equation}\label{eq:delta}
							\delta (x,y) = \begin{cases}
								1, & \text{if } x = y; \\
								0, & \text{if } x \neq y.
							\end{cases}
						\end{equation}
						
						\item \emph{\textbf{Normalized Mutual Information (NMI)}}: This measures the similarity between predicted and true labels:
						\begin{equation}
							NMI(P,Q) = \frac{MI(P,Q)}{\sqrt{H(P)H(Q)}}
						\end{equation}
						where $P$ and $Q$ are two random variables, $MI(P,Q)$ is the mutual information between $P$ and $Q$, and $H(P)$ is the entropy of $P$.
						
						\item \emph{\textbf{Purity}}: This assesses the proportion of correctly clustered samples:
						\begin{equation}
							\text{Purity} = \frac{n_1}{n_1 + n_2}
						\end{equation}
						where $n_1$ is the number of sample pairs correctly clustered, and $n_2$ is the number of incorrectly clustered pairs.
						
						All metrics range from 0 to 1, with higher values indicating better performance.
					\end{enumerate}
					
					\subsection{Parameter Setting}
					
					In the FKMWC method, there is only one hyperparameter, $\lambda$, which is associated with the scale of the input distances and the number of samples. The optimal values of $\lambda$ for our model vary based on the distance metric and dataset used:
					
					\begin{itemize}
						\item When using \textit{Euclidean distance}:
						\begin{itemize}
							\item $\lambda = 1$ to $10$ for the AR, JAFFE, ORL, UMIST, and Yaleface datasets.
							\item $\lambda = 10$ to $20$ for the MSRC\_V2 and USPS datasets.
						\end{itemize}
						\item When using the \textit{K-nearest neighbor distance}:
						\begin{itemize}
							\item $\lambda = 1$ to $10$ for the JAFFE and USPS datasets.
							\item $\lambda = 10$ to $20$ for the MSRC\_V2 dataset.
							\item $\lambda = 20$ to $30$ for the ORL dataset.
							\item $\lambda = 30$ to $40$ for the UMIST dataset.
							\item $\lambda = 50$ to $70$ for the Yaleface dataset.
						\end{itemize}
					\end{itemize}

					\subsection{Evaluation on Clustering Performance}
					
					Table~\ref{res_acc},~\ref{res_nmi}, and~\ref{res_pur} present the performance metrics of FKMWC and seven competing methods, including traditional $K$-Means, Fuzzy $K$-Means, and graph-based clustering algorithms. In cases of varying experimental results, each experiment was repeated 10 times, and the average was taken as the final result. Our variant, Ours\_d, which employs squared Euclidean distance in FKMWC, consistently outperforms $K$-Means++ (also based on Euclidean distance) across all seven benchmark datasets, demonstrating both superior and more consistent performance. In comparison to RKM, Ours\_d shows higher performance on the JAFFE, MSRC\_V2, ORL, UMIST, and USPS datasets. Similarly, Ours\_d surpasses MSFCM in most datasets. Additionally, Ours\_kd, utilizing the K-nearest neighbor distance, outperforms all other comparative algorithms, including Ksum\_x, which also employs K-nearest neighbor distance, as well as other graph-based algorithms like Ksum and ULGE. Ours\_ker, which employs kernel distance in FKMWC, due to its calculation of the distance in the kernel space, demonstrates good applicability for linearly inseparable datasets, and achieves higher results than squared Euclidean distance on six datasets. It only achieves similar results as Euclidean distance on the MSRC\_V2 dataset.

					Our proposed model, which is a novel framework for fuzzy \textit{K}-Means clustering, differs from traditional \textit{K}-Means and Fuzzy \textit{K}-Means algorithms. It uniquely allows for the flexible selection of different distance metrics based on the dataset characteristics, leading to enhanced clustering outcomes. As a result, our model exhibits greater versatility and adaptability in various clustering scenarios.

					\subsection{Parameter Sensitivity Analysis}
					
					There is only one hyperparameter, $\lambda$, in the FKMWC algorithm. When $\lambda=0$, the model described in equation~(\ref{fkmwc}) minimizes only if each sample is assigned to the nearest cluster center, which is equivalent to the classical \textit{K}-Means clustering. Figure~\ref{result_lambda} presents the ACC (Accuracy), NMI (Normalized Mutual Information), and Purity results for the FKMWC algorithm under two different distance measures: squared Euclidean distance (denoted as Ours\_d) and K-nearest neighbors distance (denoted as Ours\_kd), as the parameter $r$ varies. It is observed that for a specific dataset, the performance of the model remains relatively stable within a certain range of $\lambda$ values.

					\subsection{Convergence Analysis}
					
					The solution of FKMWC employs adaptive gradient descent. After each gradient descent step, we compute the value of the objective function~(\ref{fkmwc}). Let $\textrm{obj}_t$ denote the objective value at the $t$-th iteration. The convergence criterion of the algorithm is defined as follows: the change in the objective function value before and after one iteration should be less than $10^{-3}$, i.e.,
					$$
					\|\textrm{obj}_t - \textrm{obj}_{t-1}\|_2 \leq 10^{-3}.
					$$
					Figure~\ref{result_iter} records the variation of clustering metrics (ACC, NMI, Purity) with the number of iterations. It is observed that on all datasets, the FKMWC algorithm can reach a stable value within 2000 iterations.

					\section{Conclusion}
					
					This paper proposes an algorithm framework for fuzzy $K$-Means clustering that completely eliminates cluster centroids. It does not require random selection of initial cluster centroids nor does it need iterative updates of cluster centroids. Instead, it directly calculates the fuzzy membership matrix from distance information between samples by optimizing the objective function. The paper proves that this model is completely equivalent to the classic fuzzy $K$-Means clustering algorithm. Additionally, it not only addresses the significant impact of initial cluster centroids on the performance of traditional fuzzy $K$-Means clustering but also enhances the algorithm's robustness to noise. This model can also serve as a framework to easily extend to other algorithms by inputting different distance matrices; for instance, inputting a kernel distance would make the model equivalent to kernel fuzzy $K$-Means clustering. Experiments on seven benchmark datasets demonstrate the superior performance of the proposed algorithm.

\ifCLASSOPTIONcompsoc
\else
\fi
\ifCLASSOPTIONcaptionsoff
  \newpage
\fi

{\small
\bibliographystyle{IEEEtran}
\bibliography{egbib}

\begin{thebibliography}{10}
\providecommand{\url}[1]{#1}
\csname url@samestyle\endcsname
\providecommand{\newblock}{\relax}
\providecommand{\bibinfo}[2]{#2}
\providecommand{\BIBentrySTDinterwordspacing}{\spaceskip=0pt\relax}
\providecommand{\BIBentryALTinterwordstretchfactor}{4}
\providecommand{\BIBentryALTinterwordspacing}{\spaceskip=\fontdimen2\font plus
\BIBentryALTinterwordstretchfactor\fontdimen3\font minus
  \fontdimen4\font\relax}
\providecommand{\BIBforeignlanguage}[2]{{%
\expandafter\ifx\csname l@#1\endcsname\relax
\typeout{** WARNING: IEEEtran.bst: No hyphenation pattern has been}%
\typeout{** loaded for the language `#1'. Using the pattern for}%
\typeout{** the default language instead.}%
\else
\language=\csname l@#1\endcsname
\fi
#2}}
\providecommand{\BIBdecl}{\relax}
\BIBdecl

\bibitem{NgJW01}
A.~Y. Ng, M.~I. Jordan, and Y.~Weiss, ``On spectral clustering: Analysis and an
  algorithm,'' in \emph{{NIPS}}, 2001, pp. 849--856.

\bibitem{YangGXYG22}
H.~Yang, Q.~Gao, W.~Xia, M.~Yang, and X.~Gao, ``Multiview spectral clustering
  with bipartite graph,'' \emph{{IEEE} Trans. Image Process.}, vol.~31, pp.
  3591--3605, 2022.

\bibitem{MacQueen1967}
J.~MacQueen \emph{et~al.}, ``Some methods for classification and analysis of
  multivariate observations,'' in \emph{Proceedings of the fifth Berkeley
  symposium on mathematical statistics and probabilit}, vol.~1, no.~14.\hskip
  1em plus 0.5em minus 0.4em\relax Oakland, CA, USA, 1967, pp. 281--297.

\bibitem{LuXWGYG24}
H.~Lu, H.~Xu, Q.~Wang, Q.~Gao, M.~Yang, and X.~Gao, ``Efficient multi-view
  k-means for image clustering,'' \emph{{IEEE} Trans. Image Process.}, vol.~33,
  pp. 273--284, 2024.

\bibitem{NieLWL23}
F.~Nie, Z.~Li, R.~Wang, and X.~Li, ``An effective and efficient algorithm for
  k-means clustering with new formulation,'' \emph{{IEEE} Trans. Knowl. Data
  Eng.}, vol.~35, no.~4, pp. 3433--3443, 2023.

\bibitem{KimLLL05}
D.~Kim, K.~Y. Lee, D.~Lee, and K.~H. Lee, ``Evaluation of the performance of
  clustering algorithms in kernel-induced feature space,'' \emph{Pattern
  Recognit.}, vol.~38, no.~4, pp. 607--611, 2005.

\bibitem{WangLLNL22}
R.~Wang, J.~Lu, Y.~Lu, F.~Nie, and X.~Li, ``Discrete and parameter-free
  multiple kernel k-means,'' \emph{{IEEE} Trans. Image Process.}, vol.~31, pp.
  2796--2808, 2022.

\bibitem{RenS021}
Z.~Ren, Q.~Sun, and D.~Wei, ``Multiple kernel clustering with kernel k-means
  coupled graph tensor learning,'' in \emph{{AAAI}}, 2021, pp. 9411--9418.

\bibitem{DunnFCM1973}
J.~C. Dunn, ``A fuzzy relative of the isodata process and its use in detecting
  compact well-separated clusters,'' \emph{Journal of Cybernetics}, vol.~3,
  no.~3, pp. 32--57, 1973.

\bibitem{XuHXN16}
J.~Xu, J.~Han, K.~Xiong, and F.~Nie, ``Robust and sparse fuzzy k-means
  clustering,'' in \emph{{IJCAI}}, 2016, pp. 2224--2230.

\bibitem{NieZWLL22}
F.~Nie, X.~Zhao, R.~Wang, X.~Li, and Z.~Li, ``Fuzzy k-means clustering with
  discriminative embedding,'' \emph{{IEEE} Trans. Knowl. Data Eng.}, vol.~34,
  no.~3, pp. 1221--1230, 2022.

\bibitem{WangZNL23}
J.~Wang, X.~Zhang, F.~Nie, and X.~Li, ``Enhanced robust fuzzy k-means
  clustering joint l0-norm constraint,'' \emph{Neurocomputing}, vol. 561, p.
  126842, 2023.

\bibitem{pami/ShiM00}
J.~Shi and J.~Malik, ``Normalized cuts and image segmentation,'' \emph{{IEEE}
  Trans. Pattern Anal. Mach. Intell.}, vol.~22, no.~8, pp. 888--905, 2000.

\bibitem{ArthurV07}
D.~Arthur and S.~Vassilvitskii, ``k-means++: the advantages of careful
  seeding,'' in \emph{{ACM-SIAM} {SODA}}, 2007, pp. 1027--1035.

\bibitem{RenHC22}
J.~Ren, K.~Hua, and Y.~Cao, ``Global optimal k-medoids clustering of one
  million samples,'' in \emph{NeurIPS}, 2022.

\bibitem{NewlingF17}
J.~Newling and F.~Fleuret, ``K-medoids for k-means seeding,'' in
  \emph{NeurIPS}, 2017, pp. 5195--5203.

\bibitem{ZhouCCZL16}
J.~Zhou, L.~Chen, C.~L.~P. Chen, Y.~Zhang, and H.~Li, ``Fuzzy clustering with
  the entropy of attribute weights,'' \emph{Neurocomputing}, vol. 198, pp.
  125--134, 2016.

\bibitem{LiNCH08}
M.~J. Li, M.~K. Ng, Y.~Cheung, and J.~Z. Huang, ``Agglomerative fuzzy k-means
  clustering algorithm with selection of number of clusters,'' \emph{{IEEE}
  Trans. Knowl. Data Eng.}, vol.~20, no.~11, pp. 1519--1534, 2008.

\bibitem{tang2019possibilistic}
Y.~Tang, X.~Hu, W.~Pedrycz, and X.~Song, ``Possibilistic fuzzy clustering with
  high-density viewpoint,'' \emph{Neurocomputing}, vol. 329, pp. 407--423,
  2019.

\bibitem{DingH05}
C.~H.~Q. Ding and X.~He, ``On the equivalence of nonnegative matrix
  factorization and spectral clustering,'' in \emph{{SIAM} {SDM}}, 2005, pp.
  606--610.

\bibitem{DhillonGK04}
I.~S. Dhillon, Y.~Guan, and B.~Kulis, ``Kernel k-means: spectral clustering and
  normalized cuts,'' in \emph{{ACM} {SIGKDD}}, 2004, pp. 551--556.

\bibitem{PeiC00023}
S.~Pei, H.~Chen, F.~Nie, R.~Wang, and X.~Li, ``Centerless clustering,''
  \emph{{IEEE} Trans. Pattern Anal. Mach. Intell.}, vol.~45, no.~1, pp.
  167--181, 2023.

\bibitem{Bezdek81}
J.~C. Bezdek, \emph{Pattern Recognition with Fuzzy Objective Function
  Algorithms}.\hskip 1em plus 0.5em minus 0.4em\relax Springer, 1981.

\bibitem{5584538}
Y.~Namkoong, G.~Heo, and Y.~W. Woo, ``An extension of possibilistic fuzzy
  c-means with regularization,'' in \emph{International Conference on Fuzzy
  Systems}, 2010, pp. 1--6.

\bibitem{LuGWYX23}
H.~Lu, Q.~Gao, Q.~Wang, M.~Yang, and W.~Xia, ``Centerless multi-view k-means
  based on the adjacency matrix,'' in \emph{{AAAI}}, 2023, pp. 8949--8956.

\bibitem{AleixAR1998}
A.~Martinez and R.~Benavente,
  \emph{\BIBforeignlanguage{Indefinido/desconocido}{The AR Face Database: CVC
  Technical Report, 24}}, Jan. 1998.

\bibitem{670949}
M.~Lyons, S.~Akamatsu, M.~Kamachi, and J.~Gyoba, ``Coding facial expressions
  with gabor wavelets,'' in \emph{Proceedings Third IEEE International
  Conference on Automatic Face and Gesture Recognition}, 1998, pp. 200--205.

\bibitem{WinnJ05}
J.~M. Winn and N.~Jojic, ``{LOCUS:} learning object classes with unsupervised
  segmentation,'' in \emph{{ICCV}}, vol.~1, 2005, pp. 756--763.

\bibitem{Cai2010UnsupervisedFS}
D.~Cai, C.~Zhang, and X.~He, ``Unsupervised feature selection for multi-cluster
  data,'' in \emph{{ACM} {SIGKDD}}, 2010, pp. 333--342.

\bibitem{6565365}
C.~Hou, F.~Nie, X.~Li, D.~Yi, and Y.~Wu, ``Joint embedding learning and sparse
  regression: A framework for unsupervised feature selection,'' \emph{IEEE
  Transactions on Cybernetics}, vol.~44, no.~6, pp. 793--804, 2014.

\bibitem{291440}
J.~Hull, ``A database for handwritten text recognition research,'' \emph{IEEE
  Transactions on Pattern Analysis and Machine Intelligence}, vol.~16, no.~5,
  pp. 550--554, 1994.

\bibitem{598228}
P.~Belhumeur, J.~Hespanha, and D.~Kriegman, ``Eigenfaces vs. fisherfaces:
  recognition using class specific linear projection,'' \emph{IEEE Transactions
  on Pattern Analysis and Machine Intelligence}, vol.~19, no.~7, pp. 711--720,
  1997.

\bibitem{Ksum}
S.~Pei, H.~Chen, F.~Nie, R.~Wang, and X.~Li, ``Centerless clustering,''
  \emph{IEEE Transactions on Pattern Analysis and Machine Intelligence},
  vol.~45, no.~1, pp. 167--181, 2023.

\bibitem{ijcai2019p414}
W.~Lin, Z.~He, and M.~Xiao, ``Balanced clustering: A uniform model and fast
  algorithm,'' in \emph{{IJCAI-19}}, 7 2019, pp. 2987--2993.

\bibitem{9444882}
F.~Nie, J.~Xue, D.~Wu, R.~Wang, H.~Li, and X.~Li, ``Coordinate descent method
  for $k$k-means,'' \emph{IEEE Transactions on Pattern Analysis and Machine
  Intelligence}, vol.~44, no.~5, pp. 2371--2385, 2022.

\bibitem{ZhouLZP21}
S.~Zhou, D.~Li, Z.~Zhang, and R.~Ping, ``A new membership scaling fuzzy c-means
  clustering algorithm,'' \emph{{IEEE} Trans. Fuzzy Syst.}, vol.~29, no.~9, pp.
  2810--2818, 2021.

\bibitem{NieZL17}
F.~Nie, W.~Zhu, and X.~Li, ``Unsupervised large graph embedding,'' in
  \emph{Proceedings of the Thirty-First {AAAI} Conference on Artificial
  Intelligence, February 4-9, 2017, San Francisco, California, {USA}}, S.~Singh
  and S.~Markovitch, Eds.\hskip 1em plus 0.5em minus 0.4em\relax {AAAI} Press,
  2017, pp. 2422--2428.

\end{thebibliography}
}

\end{document}